\documentclass{article} 
\usepackage{iclr2019_conference,times}

\usepackage{amsmath,amsfonts,bm}

\def\eqref#1{equation~\ref{#1}}

\def\1{\bm{1}}

\def\vv{{\bm{v}}}

\def\vx{{\bm{x}}}

\DeclareMathAlphabet{\mathsfit}{\encodingdefault}{\sfdefault}{m}{sl}
\SetMathAlphabet{\mathsfit}{bold}{\encodingdefault}{\sfdefault}{bx}{n}

\usepackage{hyperref}
\usepackage{url}
\usepackage{changes} 
\usepackage{tikz} 
\usetikzlibrary{arrows} 

\usepackage{graphicx}  
\usepackage{subcaption}  

\title{ANALYSIS OF CONFIDENT-CLASSIFIERS FOR OUT-OF-DISTRIBUTION DETECTION}

\author{Sachin Vernekar\textsuperscript{1,}\thanks{Equal contribution.},\hspace{0.3em} Ashish Gaurav\textsuperscript{1,}\footnotemark[1],\hspace{0.3em} Taylor Denouden\textsuperscript{1},\hspace{0.3em} Buu~Phan\textsuperscript{2}\\
Department of Computer Science\textsuperscript{1}\\
University of Waterloo\\
\texttt{\{sverneka, a5gaurav, tadenoud, btphan\}@uwaterloo.ca} \\
\And
Vahdat~Abdelzad\textsuperscript{2},\hspace{0.3em} Rick Salay\textsuperscript{2},\hspace{0.3em} Krzysztof~Czarnecki\textsuperscript{2} \\
Department of Electrical and Computer Engineering\textsuperscript{2} \\
University of Waterloo \\
\texttt{\{vabdelza, rsalay, kczarnec\}@gsd.uwaterloo.ca} \\
}

\iclrfinalcopy 
\begin{document}

\maketitle

\begin{abstract}
Discriminatively trained neural classifiers can be trusted, only when the input data comes from the training distribution (in-distribution). Therefore, detecting out-of-distribution (OOD) samples is very important to avoid classification errors. In the context of OOD detection for image classification, one of the recent approaches proposes training a classifier called ``confident-classifier'' by minimizing the standard cross-entropy loss on in-distribution samples and minimizing the KL divergence between the predictive distribution of OOD samples in the low-density regions of in-distribution and the uniform distribution (maximizing the entropy of the outputs). Thus, the samples could be detected as OOD if they have low confidence or high entropy. In this paper, we analyze this setting both theoretically and experimentally. We conclude that the resulting confident-classifier still yields arbitrarily high confidence for OOD samples far away from the in-distribution. We instead suggest training a classifier by adding an explicit ``reject'' class for OOD samples.
\end{abstract}

\section{Introduction}
\begin{figure*}[ht!]
    \centering
    \begin{subfigure}[t]{0.49\textwidth}
        \centering
        \includegraphics[height=1.5in]{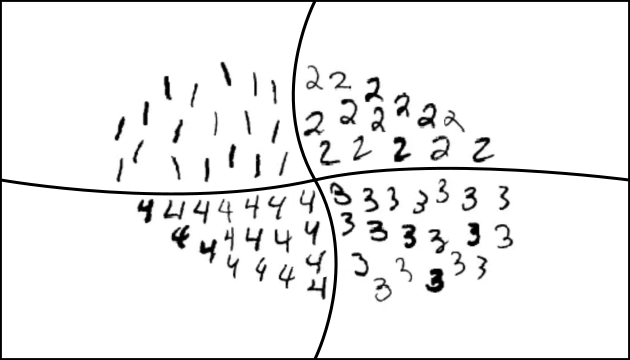}
        \caption{}
        \label{fig:plot1}
    \end{subfigure}%
    ~ 
    \begin{subfigure}[t]{0.49\textwidth}
        \centering
        \includegraphics[height=1.5in]{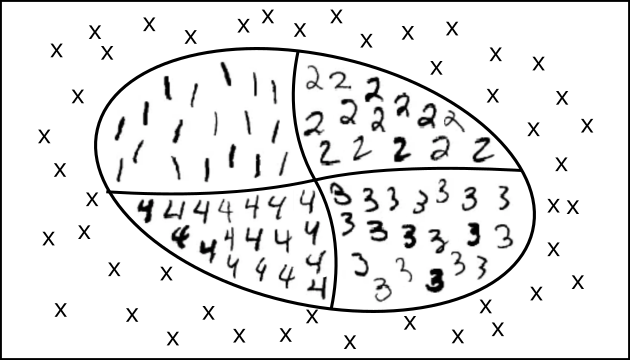}        
        \caption{}
        \label{fig:plot2}
    \end{subfigure}
    \caption{Figure shows how the decision boundaries would change and become more bounded when a typical classifier is trained with an auxiliary (``reject'') class containing OOD samples. (a) The unbounded decision boundaries of a typical 4-class classifier. (b) A 5-class classifier trained with outlier samples `x' forming the fifth (``reject'') class, that are close to in-distribution resulting in bounded decision boundaries.}
    \label{fig:problem_scatter}
\end{figure*}

Discriminatively trained deep neural networks have achieved state of the art results in many classification tasks such as speech recognition, image classification, and object detection. This has resulted in deployment of these models in real life applications where safety is paramount (e.g. autonomous driving). However, recent progress has shown that deep neural network (DNN) classifiers make overconfident predictions even when the input does not belong to any of the known classes (\cite{overConfidence}). This follows from the design of DNN classifiers that are optimized over in-distribution data without the knowledge of OOD data. The resulting decision boundaries are typically ``unbounded/open'' as shown in Figure \ref{fig:plot1} resulting in over-generalization (\cite{denoising}, \cite{openset}).

There have been many approaches proposed to address this problem under the umbrella of OOD detection. Some approaches propose to build a separate model for OOD detection (
\cite{aaeOOD}, \cite{saferClassification}, \cite{reconstructionAutoencoder}). Others propose to build OOD detection as a part of standard classifier training (\cite{trainingConfidenceCalibration}, \cite{confidenceNew}, \cite{simpleUnifiedApproach}, \cite{relu}, \cite{outlierExposure}). Our focus in this paper is on one of the latter approaches.


\cite{trainingConfidenceCalibration} propose to explicitly train a classifier using the OOD samples generated by a GAN (\cite{gan}). They empirically try to show that, for effective OOD detection, the generated OOD samples should follow and be close to the low-density boundaries of in-distribution, and the proposed GAN training indeed tries to do that. A multi-class softmax DNN classifier is trained with in-distribution samples to minimize the standard cross-entropy loss (minimizing the output entropy) and the generated OOD samples are trained with to minimize a KL loss that forces the classifier's predictive distribution to follow a uniform one (maximizing the output entropy). The resulting classifier is called a ``confident-classifier''. One can then classify a sample being in or out-of distribution based on the maximum prediction probability or the entropy of the output. \cite{confidenceNew} also follow a similar approach with slight modifications.

\textbf{Contribution.} One of the key assumptions in \cite{trainingConfidenceCalibration} and \cite{confidenceNew} is that the effect of maximizing the entropy for OOD samples close to the low-density boundaries of in-distribution might also propagate to samples that are far away from in-distribution. This training is expected to result in ``bounded/closed'' regions in input space with lower entropy over the in-distribution, and the rest of the region (corresponding to OOD), with higher entropy. The ideal decision boundary in such a scenario would be as shown in Figure \ref{fig:plot2}. We demonstrate with simple toy experiments on low-dimensional synthetic data that even though such a solution is possible, the proposed training algorithm is unlikely to reach it. We provide a theoretical argument for the same for ReLU networks (network with ReLU activation units) that was indeed used in \cite{trainingConfidenceCalibration}. Assuming training with OOD samples close to the in-distribution boundary, we find that having an explicit reject class for OOD samples results in a solution close to the one depicted in Figure \ref{fig:plot2}. We give intuitive arguments for the same. This forms the core contribution of our paper.

Moreover, with toy experiments on low-dimensional synthetic data, we analyze if GAN can indeed
produce samples that can follow the low-density boundaries of in-distribution. We find that, even though
GAN produces samples close to the low-density boundaries of in-distribution, it is unable to cover the
whole boundary, thus resulting in sub-optimal OOD detector when trained on such samples.

\section{Background}

\cite{trainingConfidenceCalibration} propose a joint training of GAN and a classifier based on the following objective:
\begin{align}
        \min_G \max_D \min_\theta &\underbrace{\mathbb{E}_{P_{in}(\hat{x},\hat{y})}[-\log P_\theta (y=\hat{y}|\hat{x})]}_{\text{(a)}} + \beta \underbrace{\mathbb{E}_{P_{G}(x)}[KL(\mathcal{U}(y)|| P_\theta(y|x))]}_\text{(b)} \nonumber \\
        &+ \underbrace{\mathbb{E}_{P_{in}(x)}[\log D(x)] + \mathbb{E}_{P_G(x)}[\log(1-D(x))]\label{(1)}}_\text{(c)}
\end{align}

where (b)+(c) is the modified GAN loss and (a)+(b) is the classifier loss ($\theta$ is the classifier's parameter) called the confidence loss. The difference from the regular GAN objective is the additional KL loss in (\ref{(1)}), which when combined with the original loss, forces the generator to generate samples in the low-density boundaries of the in-distribution ($P_{in}(x)$) space. $\beta$ is a hyper-parameter that controls how close the OOD samples are to the in-distribution boundary. For the classifier, the KL loss pushes the OOD samples generated by GAN to produce a uniform distribution at the output, and therefore have higher entropy. This enables one to detect OOD samples based on the entropy or the confidence at the output of the classifier.

\section{Why minimizing confidence loss is insufficient for OOD detection}

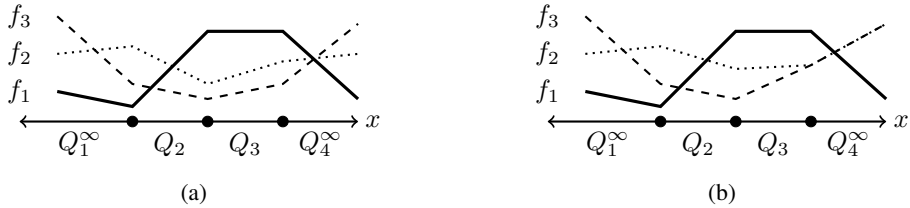
\begin{figure*}[ht!]
    \centering
    \begin{subfigure}[t]{0.49\textwidth}
        \centering
        \begin{tikzpicture}
        \draw[<->,thick] (-0.5,0) to (4,0);
        \filldraw (1,0) circle (2pt);
        \draw (0.3,-0.3)node[draw=none] {$Q_1^\infty$};
        \filldraw (2,0) circle (2pt);
        \draw (1.5,-0.3)node[draw=none] {$Q_2$};
        \filldraw (3,0) circle (2pt);
        \draw (2.5,-0.3)node[draw=none] {$Q_3$};
        \draw (3.5,-0.3)node[draw=none] {$Q_4^\infty$};
        \draw[-, dotted, thick] (0,0.9) to (1,1) to (2,0.5) to (3, 0.8) to (4, 0.9);
        \draw (-0.5, 0.9) node[draw=none]{$f_2$};
        \draw[-, dashed, thick] (0,1.4) to (1,0.5) to (2,0.3) to (3, 0.5) to (4, 1.3);
        \draw (-0.5, 1.4) node[draw=none]{$f_3$};
        \draw[-, very thick] (0,0.4) to (1,0.2) to (2,1.2) to (3, 1.2) to (4, 0.3);
        \draw (-0.5, 0.4) node[draw=none]{$f_1$};
        \draw (4.2,0) node[draw=none]{$x$};
        \end{tikzpicture}
        \caption{}
        \label{fig:hein_ex_1}
    \end{subfigure}%
    ~ 
    \begin{subfigure}[t]{0.49\textwidth}
        \centering
        \begin{tikzpicture}
        \draw[<->,thick] (-0.5,0) to (4,0);
        \filldraw (1,0) circle (2pt);
        \draw (0.3,-0.3)node[draw=none] {$Q_1^\infty$};
        \filldraw (2,0) circle (2pt);
        \draw (1.5,-0.3)node[draw=none] {$Q_2$};
        \filldraw (3,0) circle (2pt);
        \draw (2.5,-0.3)node[draw=none] {$Q_3$};
        \draw (3.5,-0.3)node[draw=none] {$Q_4^\infty$};
        \draw[-, dotted, thick] (0,0.9) to (1,1) to (2,0.7) to (3, 0.75) to (4, 1.3);
        \draw (-0.5, 0.9) node[draw=none]{$f_2$};
        \draw[-, dashed, thick] (0,1.4) to (1,0.5) to (2,0.3) to (3, 0.75) to (4, 1.3);
        \draw (-0.5, 1.4) node[draw=none]{$f_3$};
        \draw[-, very thick] (0,0.4) to (1,0.2) to (2,1.2) to (3, 1.2) to (4, 0.3);
        \draw (-0.5, 0.4) node[draw=none]{$f_1$};
        \draw (4.2,0) node[draw=none]{$x$};
        \end{tikzpicture}
        \caption{}
        \label{fig:hein_ex_2}
    \end{subfigure}
    \caption{$f_k$'s and $Q_r$'s for an example 3-class ReLU classifier where the input $x \in \mathbb{R}$. $Q_1^\infty$ and $Q_4^\infty$ are infinity polytopes. (a) For sufficiently large (small) $x$, there is a unique $k^* = 1$ in $Q_4^\infty$ ($k^* = 1$ in $Q_1^\infty$). (b) For sufficiently large $x$, there are multiple $k^*$'s in $Q_4^\infty$ ($k^* = \{2, 3\}$). For sufficiently small $x$, there is a unique $k^* = 3$ in $Q_1^\infty$.}
    \label{fig:hein_demonstrate}
\end{figure*}

Let $f: \mathbb{R}^d \rightarrow \mathbb{R}^K$ be the neural network function that maps input in $\mathbb{R}^d$ to $K$ output classes (input to the softmax layer). Let $f_k: \mathbb{R}^d \rightarrow \mathbb{R}$ be the function that maps the input to output for a specific class $k \in \{1,2,3...K\}$.
For a neural network with affine activations (eg. ReLU, Leaky ReLU), each $f_k$ is a continuous piece-wise affine function over a finite set of polytopes, $\{Q_1,Q_2,\cdots, Q_M\}$ such that $\mathbb{R}^d = \bigcup_{l=1}^{M} Q_l$, as described in \cite{randomizedGradientFreeAttack}. This means that each $f_k$ is affine within each $Q_l$ ($l \in \{1,2,3...M\}$).  If the input space is $\mathbb{R}^d$, some of these polytopes stretch to infinity (grow without bounds). Let $Q_l^\infty \equiv Q_l$ denote these ``infinity polytopes''. The choice of the neural network structure and the weights define $f_k$'s. Figure \ref{fig:hein_ex_1} illustrates these polytopes and $f_k$'s for a simple 3-class ReLU classifier, where the input space is $\mathbb{R}$. In this example, there are 4 polytopes in which $Q_1^\infty$ and $Q_4^\infty$ stretch to infinity. 



\cite{relu} mathematically show that a ReLU classifier (with softmax output) produces arbitrarily high confidence predictions (approaching 1) far away from the training data in almost all directions on an unbounded input space. This happens over $Q_l^\infty$'s. Their results are summarized as follows.

For any $\displaystyle \vx\in \mathbb{R}^d$, there exists a $\beta_l > 0$ such that for all $\alpha_l \geq \beta_l$, $\alpha_l \displaystyle \vx \in Q_l^\infty$.  Let $f_k^l(\displaystyle \vx) = \langle\vv_k^l, \displaystyle \vx\rangle+a_k^l$ be the piece-wise affine function for class $k$ over $Q_l$. Let $k^* = \arg\max_k \langle \vv_k^l, \displaystyle \beta_l \vx \rangle$\footnote{Note, $k^* = \arg\max_k \langle \vv_k^l, \alpha_l \vx \rangle = \arg\max_k \langle \vv_k^l, \beta_l \displaystyle \vx \rangle$, $\forall \alpha_l \geq \beta_l$. Also note, we define k* only for infinity polytopes.}. Then, as $\alpha_l \rightarrow \infty$, the confidence for input $\alpha_l \displaystyle \vx$ for class $k^*$ becomes arbitrarily high if $k^*$ is unique. i.e, 

\begin{align}
\displaystyle{\lim_{\alpha_l \to \infty}} \frac{e^{f_{k^*} (\alpha_l \displaystyle \vx)}}{\sum_{l=1}^K e^{f_{l} (\alpha_l \displaystyle \vx)}} = 1 \nonumber \\ 
\end{align}

But if there are multiple $k^*$'s, arbitrarily large confidence values cannot be obtained far away from the in-distribution in the direction of $x$. For instance, as shown in Figure \ref{fig:hein_ex_2}\footnote{Note, for $x\in \mathbb{R}$, $k^* = \arg \max_k [\mbox{slope of}f_k(\alpha_l x)]$ ($ = \arg \max_k [\mbox{negative slope of} f_k(\alpha_l x)]$) as $\alpha_l\rightarrow\infty$ ($\alpha_l\rightarrow -\infty$).}, for $Q_4^\infty$, $k^* = \{2, 3\}$ and therefore arbitrarily high confidence predictions cannot be achieved as $\alpha_l\rightarrow\infty$. But, having multiple $k^*$'s for every $Q_l^\infty$ is highly unlikely to happen, given that it is not explicitly enforced during training. Therefore, having arbitrarily high confidence values far away from the in-distribution is likely inevitable.


\textbf{Corollary.} Higher the confidence of the output, the lower is the entropy. Hence a direct corollary of \cite{relu}'s result is that the entropy of the classifier output for data far away from the in-distribution data in all directions would almost always be arbitrarily low (approaching 0) like the in-distribution samples. This makes it almost impossible to detect OOD samples based on the confidence or the entropy of the classifier outputs. Therefore, approaches in \cite{trainingConfidenceCalibration} and \cite{confidenceNew} would not be applicable. 

The above results are applicable for the case when the input space is $\mathbb{R}^d$. However, for images, the input space is $[0,1]^d$, and hence these results wouldn't be directly applicable. But note that one can achieve high confidence in regions far-off from the training data if one of the $f_k$'s is sufficiently greater than the rest. Such a case is likely possible for classifiers that are trained to minimize the loss (\ref{(1)}(b)) only on the data in the low-density regions of in-distribution. In the experiments section, we show that for such a classifier, one can find high confidence regions for OOD data without stretching to infinity.

\section{Adding an explicit ``reject'' class}
When OOD samples are generated close to the in-distribution and follow its low-density boundaries as proposed in \cite{trainingConfidenceCalibration} and \cite{confidenceNew}, we recommend adding an explicit ``reject'' class for OOD samples instead of minimizing the loss in (\ref{(1)}(b)). Let the resulting classifier be called the "reject-classifier". The intuition is as follows. The arbitrarily high confidence predictions happen in polytopes that stretch to infinity. Each of the infinity polytopes has its own class (or classes), $k^*$(or $k^*$'s) where high confidence predictions occur. If adding an explicit ``reject'' class results in $k^* = \mbox{reject-class}$ for all the infinity polytopes, the arbitrarily high confidence predictions would only happen at the reject class for OOD samples far-off from training data. We argue that the reject-classifier training might result in such a solution depending on how well the OOD samples follow the in-distribution boundary. For example, Figure. \ref{fig:plot2} indicates the resulting decision boundaries for an ideal reject-classifier. The experiments in the next section support this claim. Although we remark, we can't theoretically guarantee that reject-classifier training would necessarily result in such a solution.


\cite{trainingConfidenceCalibration} indeed experiment with adding an explicit reject class instead of using a confident-classifier, but the results are found to be worse. But this could be because the generated OOD samples don't follow the in-distribution boundaries well (refer to section \ref{OOD sample generation}). \cite{outlierExposure} and \cite{relu} also propose to train a classifier with confidence loss where OOD data is obtained from a large natural dataset or is synthetically generated by random sampling on the input space. Therefore the OOD samples here are not limited to the ones close to the in-distribution. In this case, the choice between using a confidence loss and adding a ``reject'' class is arbitrary with respect the results from section 3, although \cite{outlierExposure}'s experiments support using a confidence loss. However, such approaches are only feasible for input spaces where (approximately) representing the entire OOD region with a finite number of samples is possible. This is definitely not possible for example when the input space is $\mathbb{R}^d$.


\section{Experiments}

In all our experiments\footnote{Our code is available at \href{https://github.com/sverneka/ConfidentClassifierICLR19}{https://github.com/sverneka/ConfidentClassifierICLR19}.}, the input space is $\mathbb{R}^2$ and the in-distribution consists of 2-classes. The samples for each of these classes are generated by sampling from 2 Gaussians with identity co-variances and means (-10, 0) and (10, 0) respectively, on the Cartesian coordinates. Anything outside 3 standard deviations (Mahalanobis distance) from the in-distribution means is considered OOD. Unless otherwise specified, the neural network used is similar to the one used in \cite{trainingConfidenceCalibration}, which is a ReLU-classifier with 2 fully-connected hidden layers with 500 neurons each.

\subsection{Training classifiers on OOD samples}
We consider 2 cases with respect to how OOD samples are generated.

\begin{figure}[!htb]
\subfigure{0.32\textwidth}
  \includegraphics[width=\linewidth]{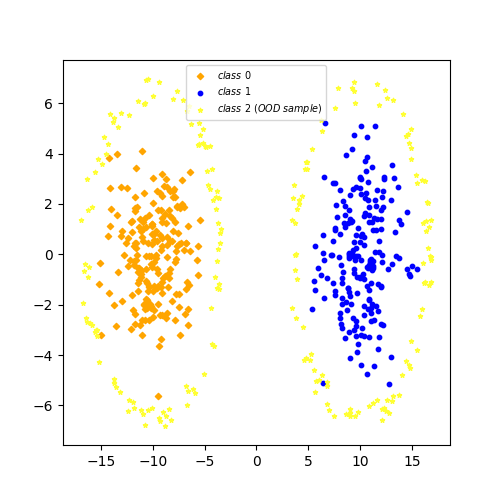}
  \caption{}\label{fig:exp1_a}
\endsubfigure\hfill
\subfigure{0.32\textwidth}
  \includegraphics[width=\linewidth]{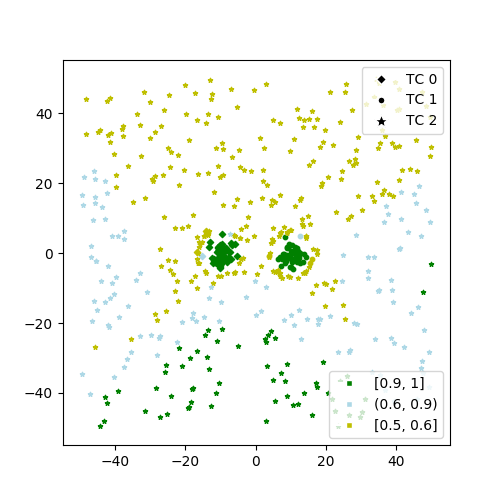}
  \caption{}\label{fig:exp1_b}
\endsubfigure\hfill
\subfigure{0.32\textwidth}%
  \includegraphics[width=\linewidth]{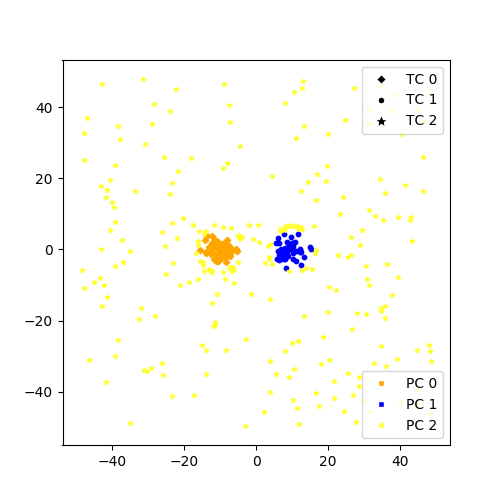}
  \caption{}\label{fig:exp1_c}
\endsubfigure
\caption{Plots for boundary OOD samples experiments. (a) Training data in 2D. (b) Maximum prediction output on test data for a confident-classifier. (c) Classification output of a classifier with a ``reject'' class on test data (TC = true class, PC = predicted class).}
\label{fig:exp1}
\end{figure}

\textbf{Boundary OOD samples.}
Following the case in \cite{trainingConfidenceCalibration}, for training, OOD samples are generated close to the in-distribution as shown in Figure \ref{fig:exp1_a}. For testing, OOD samples are uniformly sampled from a 2D box $[-50, 50]^2$ excluding the in-distribution regions.

From Figure \ref{fig:exp1_b}, we observe that the ReLU-classifier trained to optimize confidence loss results in highly confident predictions for many OOD samples far from the in-distribution data. This renders the classifier ineffective at classifying the in and out of distribution samples based on the maximum prediction score (confidence) or the entropy of the output. However, from Figure \ref{fig:exp1_c}, for a classifier trained with explicit reject class, the test OOD samples are indeed classified as OOD. This supports the aforementioned intuitions.

Note that these are not the results specific to a certain architecture of the neural network. Experiments with different hyper-parameters such as the number of hidden neurons, changing input dimensions, using sigmoid activation functions instead of ReLU lead to similar results. We remark however that for sigmoid networks, the results were not as extreme (in terms of the number of OOD samples with high-confidence) as for ReLU networks. This is understandable because sigmoid activation outputs will not produce arbitrarily large values, unlike the ReLU counterparts.

\begin{figure}[!htb]
\subfigure{0.32\textwidth}
  \includegraphics[width=\linewidth]{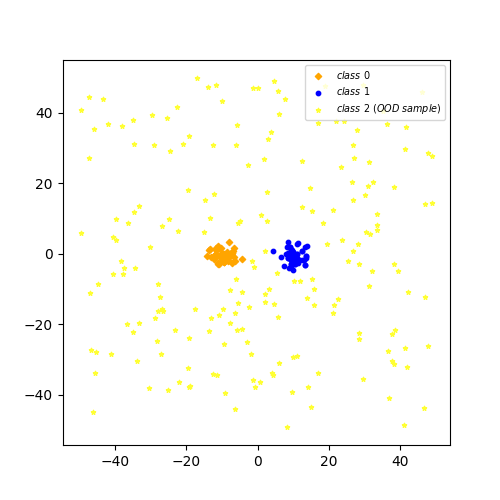}
  \caption{}\label{fig:exp2_a}
\endsubfigure\hfill
\subfigure{0.32\textwidth}
  \includegraphics[width=\linewidth]{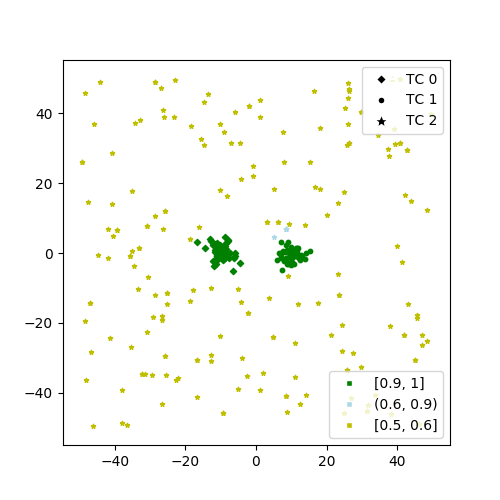}
  \caption{}\label{fig:exp2_b}
\endsubfigure\hfill
\subfigure{0.32\textwidth}%
  \includegraphics[width=\linewidth]{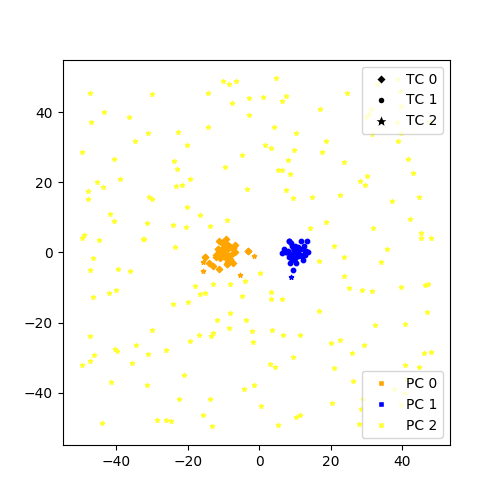}
  \caption{}\label{fig:exp2_c}
\endsubfigure
\caption{Plots for general OOD samples experiments. (a) Training data in 2D. (b) Maximum prediction output on test data for a confident-classifier. (c) Classification output of a classifier with a ``reject'' class on test data.}
\label{fig:exp2}
\end{figure}

\textbf{General OOD samples.} In this case, both train and test OOD samples are uniformly sampled from a 2D box $[-50, 50]^2$ excluding the in-distribution regions. From Figure \ref{fig:exp2}, we observe that both confidence loss and reject class based classifiers are able to distinguish in and out of distribution samples effectively. Therefore, there is no clear winner between the two.

\subsection{Generating OOD samples using GAN} \label{OOD sample generation}

\cite{trainingConfidenceCalibration} propose to generate OOD samples in the low-density regions of in-distribution by optimizing a joint GAN-classifier loss, (\ref{(1)}). With a toy experiment, they show that the generator indeed produces such samples and also these samples follow the ``boundary'' of the in-distribution data. However, in the experiment, they use a pre-trained classifier. The classifier is pre-trained to optimize the confidence loss on in-distribution and OOD samples sampled close to the in-distribution. Therefore the classifier already has the knowledge of those OOD samples. When GAN is then trained following the objective in (\ref{(1)}), GAN likely generates those OOD samples close to the in-distribution. But it is evident that this setting is not realistic as one cannot have a fully informative prior knowledge of those OOD samples if our objective is to generate them.

\begin{figure}[!htb]
\subfigure{0.32\textwidth}
  \includegraphics[width=\linewidth]{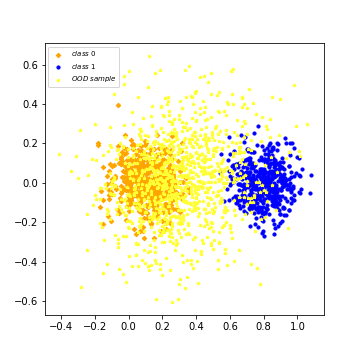}
  \caption{Epoch 100}\label{fig:exp3_a}
\endsubfigure\hfill
\subfigure{0.32\textwidth}
  \includegraphics[width=\linewidth]{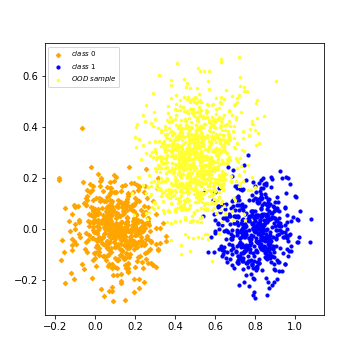}
  \caption{Epoch 500}\label{fig:exp3_b}
\endsubfigure\hfill
\subfigure{0.32\textwidth}%
  \includegraphics[width=\linewidth]{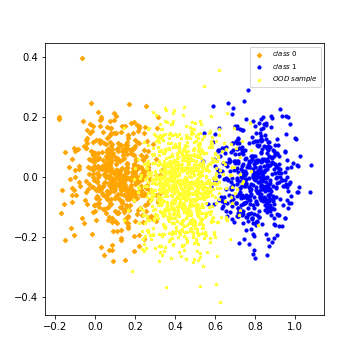}
  \caption{Epoch 1000}\label{fig:exp3_c}
\endsubfigure
\caption{Generated OOD samples using a joint training of a GAN and a confident-classifier. We observe that the generated OOD samples don't cover the entire in-distribution boundary.}
\label{fig:exp3}
\end{figure}

Therefore, we experimented by directly optimizing (\ref{(1)}) where the classifier is not pre-trained. For a 2D dataset case, as shown in Figure \ref{fig:exp3}, we find that (with much hyper-parameter tuning), even though GAN ends up producing OOD samples close to the in-distribution, it does an unsatisfactory job at producing samples that could follow the entire in-distribution boundary. Moreover, there is less diversity in the generated samples which make them ineffective at improving the classifier performance in OOD detection. Our intuition is that the loss (\ref{(1)}(b)+\ref{(1)}(c)) that forces the generator of the GAN to generate samples in the high entropy regions of the classifier doesn't necessarily enforce it to produce samples that follow the entire in-distribution boundary. The inability of GANs to generate such samples for a simple 2D dataset indicates that it would be even more difficult in higher dimensions.

\section{Conclusion}
We have shown in the paper that the confident-classifier almost always has OOD samples that produce high confidence outputs (in the contexts described earlier). We provided empirical evidence that favor using an explicit ``reject'' class instead. However, the ODD detection capabilities of a reject-classifier depend on the extent to which the generated OOD samples follow the low-density boundaries of in-distribution. We also have shown how optimizing (\ref{(1)}) doesn't produce desired OOD samples. Therefore, for the future, it would be desirable to investigate other approaches that can generate such samples.



\bibliography{iclr2019_conference}
\bibliographystyle{iclr2019_conference}

\end{document}